\documentclass{article}
\usepackage{ijcai21}

\usepackage{times}
\usepackage{soul}
\usepackage{url}
\usepackage[hidelinks]{hyperref}
\usepackage[utf8]{inputenc}
\usepackage[T1]{fontenc}
\usepackage[small]{caption}
\usepackage{graphicx}
\usepackage{natbib}
\usepackage{amsmath}
\usepackage{amsthm}
\usepackage{booktabs}
\usepackage{algorithm}
\usepackage{algorithmic}
\usepackage{todonotes}
\usepackage{caption}
\usepackage{amsfonts}
\usepackage{cleveref}

\title{Conditional Generative Models for Counterfactual Explanations}

\author{
Arnaud Van Looveren$^1$\footnote{Contact Author}\and
Janis Klaise$^1$\and
Giovanni Vacanti$^1$\And
Oliver Cobb$^1$\\
\affiliations
$^1$Seldon Technologies\\
\emails
\{avl, jk, gv, oc\}@seldon.io
}

\begin{document}

\maketitle

\begin{abstract}
  Counterfactual instances offer human-interpretable insight into the local behaviour of machine learning models. We propose a general framework to generate sparse, in-distribution counterfactual model explanations which match a desired target prediction with a conditional generative model, allowing batches of counterfactual instances to be generated with a single forward pass. The method is flexible with respect to the type of generative model used as well as the task of the underlying predictive model. This allows straightforward application of the framework to different modalities such as images, time series or tabular data as well as generative model paradigms such as GANs or autoencoders and predictive tasks like classification or regression. We illustrate the effectiveness of our method on image (CelebA), time series (ECG) and mixed-type tabular (Adult Census) data.
\end{abstract}

\section{Introduction}
Recent improvements in the predictive ability of machine learning models has lead to their increasingly widespread deployment within automated decision-making systems of real-world consequence. However, the increase in model complexity that has given rise to these improvements has simultaneously hampered our ability to understand model decision-making processes. This has motivated the design of tools and methods that analyse why models make certain decisions and not others. For example, such insight may be of crucial importance in analysing a car accident involving an autonomous driving system, checking the rationale behind a particular medical diagnosis, or providing explanation to a customer who has had a loan application denied.



\begin{figure}[ht]
    \centering
    \includegraphics[width=\columnwidth]{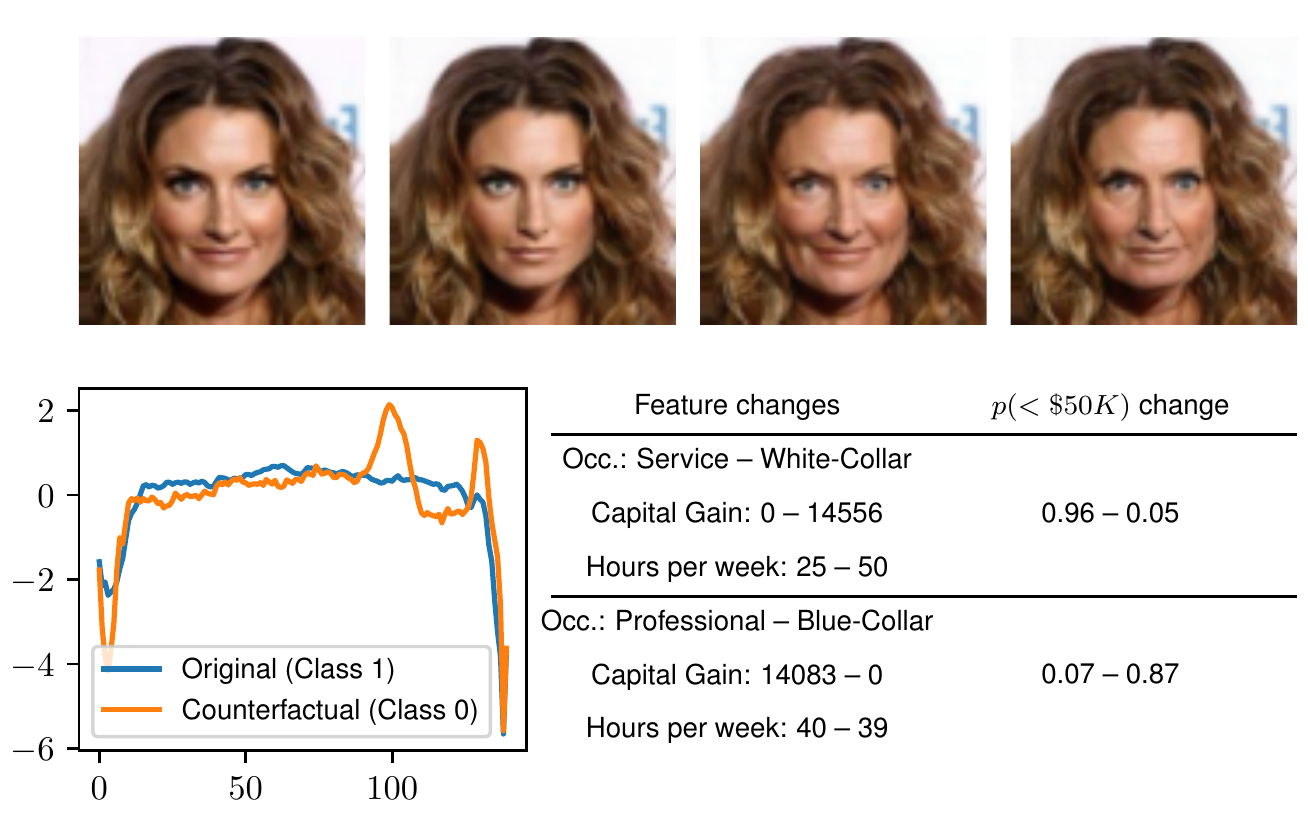}
    \caption{Counterfactual instances changing the classifier's prediction. \emph{Top row: }Images---young/smiling (original), young/non-smiling (counterfactual), old/smiling (counterfactual), old/non-smiling (counterfactual). \emph{Bottom left: } Time series---abnormal ECG (original), normal ECG (counterfactual). \emph{Bottom right: } tabular---Two instances of individuals with counterfactual feature changes flipping the classifier prediction from low income to high (top instance) and vice versa (bottom instance).}
    \label{fig:cf_all_modality}
\end{figure}

A powerful way to obtain such insight is through the analysis of \emph{counterfactual instances}. A counterfactual instance is defined as a synthetic instance for which a trained machine learning model predicts a desired output which is different from the prediction made on the original instance (\Cref{fig:cf_all_modality}). To provide a useful, plausible alternative for the original instance, the counterfactual instance should be statistically indistinguishable from real instances. We will refer to this property as \emph{in-distributionness} of the counterfactuals relative to the data distribution of the overall training data or relative to the distribution of training instances belonging to a specific class (class-conditional in-distributionness). Moreover, a counterfactual instance should be relatively close to the original instance to allow the change in the model prediction to be attributed more precisely. We will refer to this second property as \emph{sparsity}, meaning that the difference between the original and counterfactual instances should be sparse.


Most existing methods to generate counterfactuals iteratively perturb the features of the original instance during an optimization process until the target prediction is met. The perturbations are typically encouraged to be sparse and in-distribution using different loss terms. This approach, however, requires a separate optimization process for each instance to be explained, making it impractical for large amounts of instances or high-dimensional data. Here we introduce a general framework based on generative networks which allow us to create batches of counterfactual instances with a single forward pass. The generative counterfactual model is trained to predict the counterfactual perturbations or instances directly. This approach is more scalable and easily extends to different modalities. Our experiments show that the method  consistently obtains state-of-the-art results for image (CelebA~\citep{liu_celeba}), time series (ECG~\citep{baim_ecg}) and mixed type tabular data (Adult Census~\citep{uci}).

To summarize, our contributions are as follows:
\begin{itemize}
    \item A general framework to generate counterfactual examples for different data modalities.
    \item Improved quality of counterfactuals (measured by overall and class-conditional in-distributionness) due to the use of generative models.
    \item Fast generation of counterfactuals since no optimization is required at prediction time.
    \item Experiments and comparisons to baselines on image, time series and tabular datasets.
\end{itemize}

\section{Related Work}

Counterfactual instances are an alternative to feature attribution methods for explaining individual model predictions. Traditional counterfactual search methods iteratively perturb individual instances until their predictions match a specified target under a sparsity penalty \citep{wachter_cf,mothilal_dice} or apply a heuristic search procedure \citep{laugel_comparison}. These approaches can be slow and do not take the underlying data distribution into account which leads to potentially unrealistic, out-of-distribution counterfactual instances, especially on higher dimensional data such as natural images. In attempts to improve the realism of counterfactuals, \cite{vanlooveren_cfproto} use prototypes to guide the search process, \cite{liu_cfintro} leverage a pretrained conditional GAN \citep{mirza_cgan} and \cite{joshi_actionable} optimize for a perturbation in the latent space of a VAE \citep{kingma_vae}. The need for a separate optimization process for each instance remains a bottleneck that needs to be addressed in order to scale up counterfactual methods.

Conditional generative models address this issue and make it possible to generate in-distribution counterfactual instances with a single forward pass. Progress in generative models across modalities such as images \citep{karras_stylegan,brock_biggan}, time series \citep{cristobal_rcgan} or tabular data \citep{ctgan2019} can be leveraged in the counterfactual setting. \cite{mahajan_preserving} use VAEs to generate counterfactuals but require an oracle to obtain feasible instances for training purposes. Recent work by \cite{oh_bornid} is more similar to our approach but uses a conditional U-Net \citep{unet_ronneberget} as the generative model and requires the ground truth for the training data. An extensive overview of counterfactual methods by modality can be found in the survey paper by \cite{karimi_survey}.

\section{Method}
\subsection{General}

The goal is to generate sparse, in-distribution counterfactual instances $x_{\text{CF}}$ which change the prediction of model $M$ on instance $x$ from $y_{\text{M}}=M(x)$ to a target prediction $y_{\text{T}}$ with a single forward pass of a generative model $G_{\text{CF}}$ instead of solving an optimization problem at prediction time for each instance $x$. $y_{\text{T}}$ is created by applying a transformation $T$ to $y_{\text{M}}$. $T$ depends on the prediction task; for multi-class classification problems $T$ could for instance flip the predicted class from $c_{i}$ to $c_{j \neq i}$. $G_{\text{CF}}$ either generates $x_{\text{CF}}$ directly or returns a counterfactual perturbation $\delta_{\text{CF}}$ such that $x_{\text{CF}}=x+\delta_{\text{CF}}$, and is conditioned on the original instance $x$, $y_{\text{M}}$, $y_{\text{T}}$ and optionally injected noise $z$:

\begin{equation}\label{eq:cf_def}
    \begin{aligned}
        y_{\text{M}} &= M(x) \\
        y_{\text{T}} &= T(y_{M}) \\
        \delta_{\text{CF}} \: \text{or} \: x_{\text{CF}} &= G_{\text{CF}}(x, y_{\text{M}}, y_{\text{T}}, z) \\
        y_{\text{CF}} &= M(x_{\text{CF}}).
    \end{aligned}
\end{equation}

$G_{\text{CF}}$ is trained to minimize a loss $L_{G_{\text{CF}}}$ of the following form:

\begin{equation}\label{eq:generic_loss}
    \begin{aligned}
        L_{G_{\text{CF}}} &= L_{\text{M}} + L_{l_{p}} + L_{\chi} \\
        L_{\text{M}} &= w_{\text{M}} d_{\text{M}}(y_{\text{CF}}, y_{\text{T}}) \\
        L_{l_{p}} &= w_{l_{p}} d_{p}(x, x_{\text{CF}}),
    \end{aligned}
\end{equation}
where $d_{\text{M}}$ represents a divergence metric. For classification tasks we use the cross-entropy between $y_{\text{CF}}$ and $y_{\text{T}}$, but this could also be the RMSE for regression tasks. $L_{l_{p}}$ induces sparsity of the counterfactual by minimizing the $l_{p}$-metric between $x$ and $x_{\text{CF}}$ where $p$ depends on the data modality. For example $p=1$ for continuous numerical features but $p=0$ for text data since the sparsity of $x_{\text{CF}}$ can be defined as the number of tokens that have been changed. $L_{\chi}$ penalizes out-of-distribution counterfactuals, where $\chi$ represents the training data distribution. The exact form of $L_{\chi}$ depends on the type of generative model used as well as the training procedure. Note that the method does not require the ground truth of even the training instances.

\subsection{Image} \label{method:image}

The introduction of GANs enabled the generation of high-resolution \citep{karras_stylegan} and diverse \citep{brock_biggan} images. This makes GANs well suited to generating counterfactual images. The original instance $x$ is fed as the input of $G_{\text{CF}}$ instead of a random noise vector $z$. $z$ can still be injected in $G_{\text{CF}}$ to improve training. The generator is further conditioned on $y_{\text{M}}$ and $y_{\text{T}}$ and can either generate the counterfactual perturbation $\delta_{\text{CF}}$ or directly model $x_{\text{CF}}$. The task of discriminator $D$ is to distinguish real instances from the generated counterfactuals. $D$ is conditioned on $y_{\text{M}}$ for the real instances $x$ and on the target predictions $y_{\text{T}}$ for the counterfactuals $x_{\text{CF}}$. 

$L_{\chi}$ consists of the conditional GAN generator loss $L_{\text{G}}$ as well as a cycle consistency loss $L_{\text{CC}}$ \citep{zhu_cyclegan}. $L_{\text{CC}}$ requires $G_{\text{CF}}$ to map $x_{\text{CF}}$ back to $x$ and encourages only the target-specific attributes of $x_{\text{CF}}$ to change. To enforce sparse counterfactual perturbations we use the $l_{1}$-metric for $L_{l_{p}}$. The generator and discriminator losses $L_{G_{\text{CF}}}$ and $L_{D}$ are minimized in an adversarial setting, which leads to the following loss formulation:

\begin{equation}\label{eq:img_gan_loss}
    \begin{aligned}
        L_{\text{G}} &= \mathbb{E}_{x \sim \chi, z \sim p(z)}[\log(1 - D(x_{\text{CF}}, y_{\text{T}}))] \\
        L_{\text{CC}} &= \mathbb{E}_{x \sim \chi, z \sim p(z)}[\Vert x - G_{\text{CF}}(x_{\text{CF}}, y_{\text{CF}}, y_{\text{M}}, z) \Vert_{1}] \\
        L_{\chi} &= w_{G} L_{\text{G}} + w_{\text{CC}} L_{\text{CC}} \\
        L_{\text{D}} &= -\mathbb{E}_{x \sim \chi}[\log(D(x, y_{\text{M}}))] \\ & - \mathbb{E}_{x \sim \chi, z \sim p(z)}[\log(1 - D(x_{\text{CF}}, y_{\text{T}}))],
    \end{aligned}
\end{equation}
where we assume the case where $G_{\text{CF}}$ models $x_{\text{CF}}$ directly for $L_{\text{CC}}$.

\subsection{Time Series} \label{method:ts}

GANs have also proven useful for a variety of time series applications such as audio generation \citep{donahue_wavegan} or medical  data simulation \citep{cristobal_rcgan}.

We adapt the RCGAN architecture \citep{cristobal_rcgan} for our counterfactual generator $G_{\text{CF}}$ and keep the original discriminator $D$. Both the generator and discriminator networks of RCGAN consist of LSTMs \citep{hochreiter_lstm}. At each step $n$ of the sequence with total length $N$, $G_{\text{CF}}$ takes $x_{n}$, an independently sampled noise vector $z$ and the embeddings of $y_{\text{M}}$ and $y_{\text{T}}$ as inputs. $D$ is again conditioned on $y_{\text{M}}$ for real instances and $y_{\text{T}}$ for counterfactuals. This allows us to reuse the loss formulation of \eqref{eq:img_gan_loss} with only minor modifications for $L_{\text{G}}$ and $L_{\text{D}}$:

\begin{equation}\label{eq:ts_gan_loss}
    \begin{aligned}
        L_{\text{G}} &= \mathbb{E}_{x \sim \chi, z \sim p(z)}\left[\sum_{n=1}^{N}  \log(1 - D_{n}(x_{\text{CF},n}, y_{\text{T}}))\right] \\
        L_{\text{D}} &= -\mathbb{E}_{x \sim \chi}\left[\sum_{n=1}^{N} \log(D_{n}(x_{n}, y_{\text{M}}))\right] \\&- \mathbb{E}_{x \sim \chi, z \sim p(z)}\left[\sum_{n=1}^{N} \log(1 - D_{n}(x_{\text{CF},n}, y_{\text{T}}))\right],
    \end{aligned}
\end{equation}
where $D_{n}(x_{.,n}, y_{.})$ is the prediction of $D$ at step $n$ which enables more granular discriminator feedback. $L_{l_{1}}$ is used to induce sparsity.

In \cref{sec:cf_ae}, we also introduce an alternative method within the same counterfactual generation framework for time series which uses autoencoders instead of GANs to model the counterfactual instances.

\subsection{Tabular} \label{method:tabular}
Generative models for tabular data, as opposed to image or time series data, additionally require the flexibility to model relationships across heterogeneous data types. In the simplest case, a generative model will have to handle the generation of both real-valued, continuous features and categorical features. To this end, we adapt the CTGAN approach and architecture~\citep{ctgan2019}. Both the discriminator and generator are fully connected networks with residual connections, the data is represented by mode-specific normalization of continuous features and one-hot encoding of categorical features, and the generator uses Gumbel-Softmax~\citep{gumbelsoftmax2017} sampling to model categorical features. We dispense with the use of conditional vectors and training-by-sampling as for our use case the generator is already conditioned on real instances. The discriminator $D$ is again conditioned on $y_\text{M}$ for real instances and $y_\text{T}$ for counterfactuals. We use vanilla GAN losses for $L_G$ and $L_D$ and use $l_1$-metric on the data representation to induce sparsity and we don't include a cycle consistency loss. Finally, due to the presence of categorical features, we directly generate counterfactuals $x_{\text{CF}}$ instead of modelling a perturbation. This gives the following formulation of the loss:

\begin{equation}\label{eq:tabular_loss}
    \begin{aligned}
        L_{\text{G}} &= \mathbb{E}_{x \sim \chi, z \sim p(z)}[\log(1 - D(x_{\text{CF}}, y_{\text{T}}))] \\
        L_{\chi} &= w_{G} L_{\text{G}} \\
        L_{\text{D}} &= -\mathbb{E}_{x \sim \chi}[\log(D(x, y_{\text{M}})] \\ & - \mathbb{E}_{x \sim \chi, z \sim p(z)}[\log(1 - D(x_{\text{CF}}, y_{\text{T}}))].
    \end{aligned}
\end{equation}



\section{Experiments}

\subsection{Image}
\subsubsection{Dataset}

A classification model and the counterfactual generator are trained on the CelebA dataset \citep{liu_celeba} which consists of more than $200,000$ images of faces, each with $40$ attribute annotations. The images are scaled to resolution $128$x$128$ and divided into $4$ non-overlapping classes based on the presence of the \textit{smiling} and \textit{young} face attributes.

\subsubsection{Models}

The original BigGAN architecture is adjusted to serve as a counterfactual generator which returns $x_{\text{CF}}$. Since $G_{\text{CF}}$ takes $x$ directly as an input, no upsampling takes place. The class-conditional BatchNorm layers in $G_{\text{CF}}$ are conditioned on separate embedding layers for $y_{\text{M}}$ and $y_{\text{T}}$, concatenated with the skip-$z$ noise vectors. We use a channel multiplier of 24 and remove the self-attention module to reduce the memory footprint. No orthogonal regularization is applied. Similarly to BigGAN, we optimize the hinge loss version of $D$ and $G_{\text{CF}}$. The classifier is a ResNet-18 \citep{he_resnet} which achieves 81.8\% accuracy on the test set.

The loss weights $w_{\text{M}}$, $w_{l_{1}}$, $w_{\text{CC}}$ and $w_{\text{G}}$ are set to 0.5, 2, 2 and 1 respectively. $w_{\text{M}}$ can stay relatively low and still allow $G_{\text{CF}}$ to generate counterfactual instances where $y_{\text{CF}}$ matches $y_{\text{T}}$. $w_{l_{1}}$ and $w_{\text{CC}}$ are both $l_{1}$-based pixel-level loss terms and set to the same value. $w_{\text{G}}$ is set to 1 and will eventually dominate as training progresses, refining the attributes of the sparse counterfactual $x_{\text{CF}}$.

We compare our method against BIN \citep{oh_bornid} who utilize a U-Net as a counterfactual generator where the skip connections between the encoder and decoder are conditioned on the target prediction in one-hot encoded format. The encoder adopts the convolutional base and frozen weights of the ResNet classifier while the decoder consists of ResBlocks which upsample the encoding back to the original input size. The discriminator also adopts the encoder's architecture.  BIN requires the ground truth of the training instances in a cycle-consistency loss term. More details on our model and the baseline can be found in \cref{sec:image_details}.

\subsubsection{Evaluation}
\begin{figure}[ht]
    \centering
    \includegraphics[width=0.95\columnwidth]{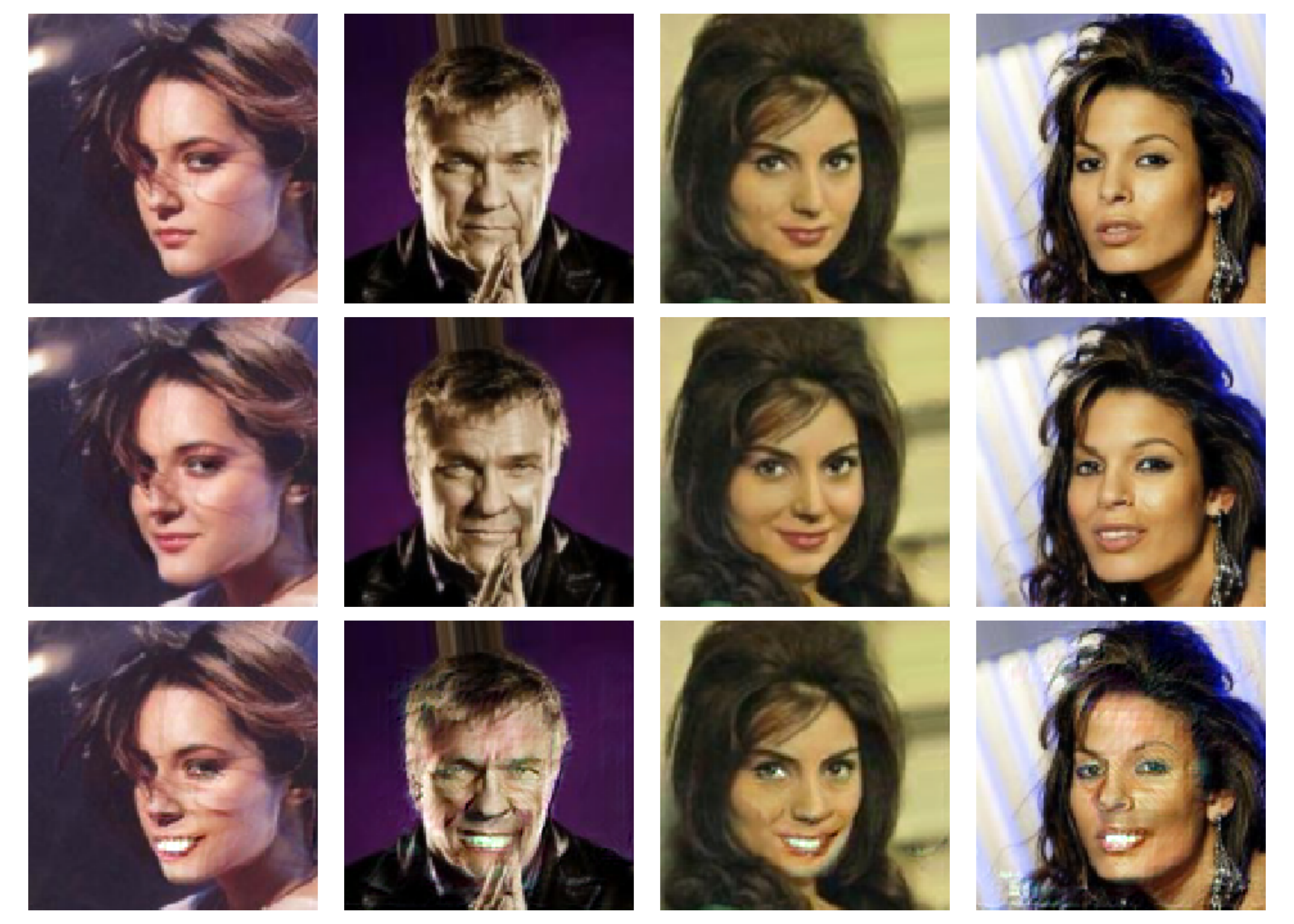}
    \caption{The top, second and third rows contain respectively the original instance, counterfactuals which introduce a smile generated by our method, and BIN \citep{oh_bornid} trying to achieve the same. Our method takes the semantics of the image into account while BIN creates the counterfactual by adding a similar smile-effect to each instance.}
    \label{fig:img_cf_unet}
\end{figure}

We measure the perceptual quality of the generated counterfactual instances via the Fr\'{e}chet Inception Distance (FID) and Inception Score (IS) metrics. For comparison we evaluate the FID and IS on 50,000 counterfactuals generated on the test set for both $G_{\text{CF}}$ and the BIN generator. \Cref{tab:img_comparison} shows that our method outperforms BIN significantly on both metrics. The difference is most noticeable in the FID score: 5.76 for our method compared to 96.56 for BIN. This can be attributed to the fact that $G_{\text{CF}}$ makes semantic changes to the image while BIN tends to apply similar transformations between classes (e.g. from \textit{non-smiling} to \textit{smiling}) regardless of the semantics of the original image, as illustrated in \Cref{fig:img_cf_unet}. Despite training for only 60,000 steps, the FID and IS values for our counterfactual instances are similar to the ones achieved by the samples of an original BigGAN model trained for 400,000 steps, a batch size of 50, with a channel multiplier of 64 and a self-attention module which reaches FID and IS values of respectively 4.54 and 3.23 \citep{schonfeld_ugan}.

\begin{figure}[ht]
    \centering
    \includegraphics[width=0.95\columnwidth]{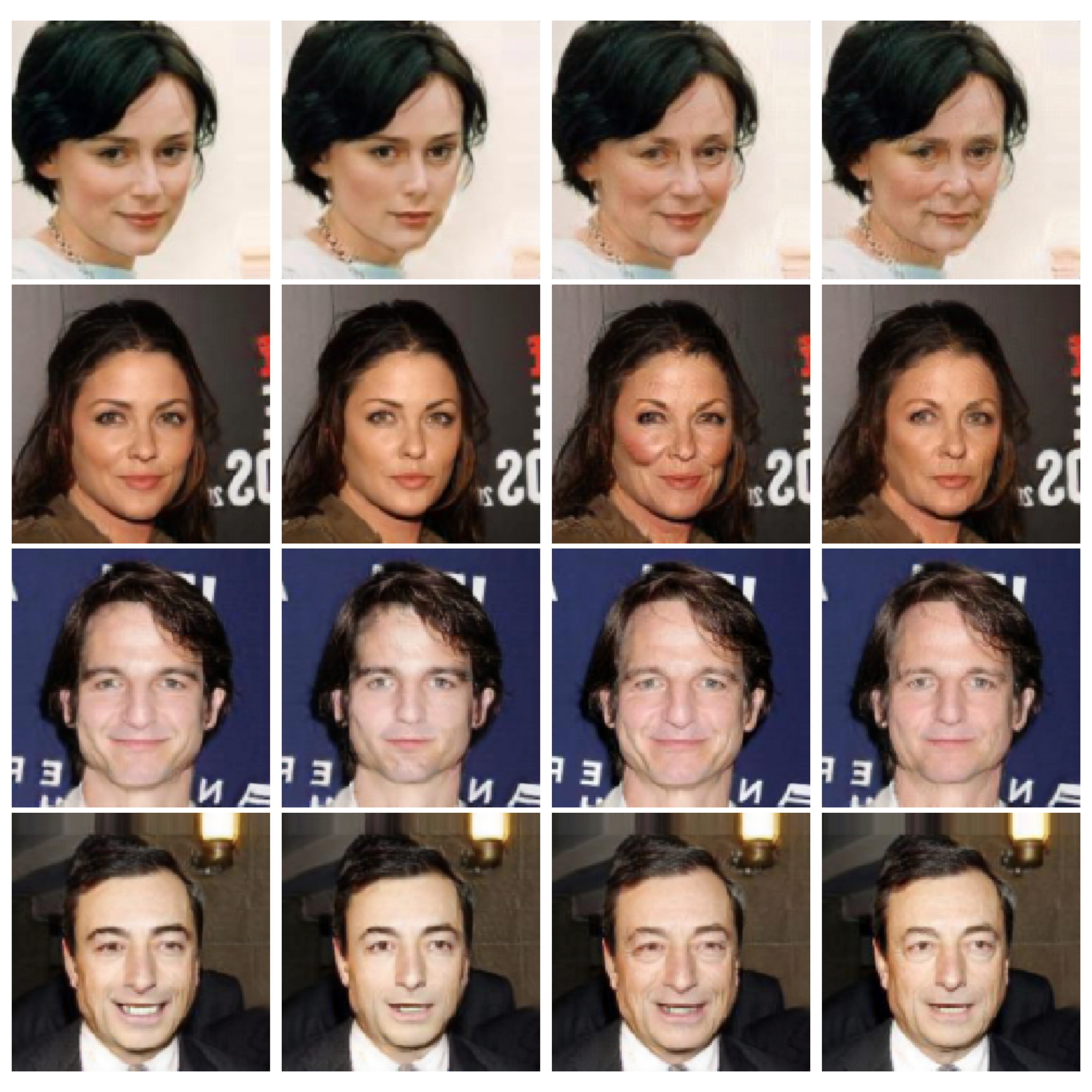}
    \caption{Counterfactual instances for each of the classes in CelebA. From left to right: young/smiling, young/non-smiling, old/smiling, old/non-smiling. The images on the diagonal are the original instances and the off-diagonal elements are counterfactuals.}
    \label{fig:img_cf}
\end{figure}

\begin{table}[ht]
\centering
\begin{tabular}{lccc}
\toprule
Method  & FID & IS & MMD$^2$ $(1e-4)$ \\ 
\midrule
BIN      &  96.56 & 2.89 $\pm$ 0.01 &  11.29  \\
Ours    & \textbf{5.76}  & \textbf{3.33 $\pm$ 0.03}  &  \textbf{1.18}  \\
\bottomrule
\end{tabular}
\caption{Image counterfactual method comparison. \emph{Lower is better} for FID and MMD$^2$ while \emph{higher is better} for IS.}
\label{tab:img_comparison}
\end{table}

We also evaluate the distance between the distributions of the test set and the counterfactual samples for both methods via the maximum mean discrepancy (MMD) \citep{gretton_mmd}. Since we are dealing with 128x128x3 images, the instances first undergo a dimensionality reduction step to an embedding dimension of 32 with a randomly initialized encoder \citep{rabanser_failingloudly}. The MMD$^2$ is then computed on the image encodings. The MMD$^2$ values for both methods shown in \Cref{tab:img_comparison} support the findings from the perceptual quality metrics and emphasize the strength of our method whose MMD$^2$ is an order of magnitude smaller than BIN.

\subsection{Time Series}

\subsubsection{Dataset}

The dataset contains 5,000 electrocardiograms (ECGs) obtained from a patient with severe congestive heart failure \citep{baim_ecg}. The ECGs have been preprocessed as follows: first each heartbeat is extracted, then each beat is made equal length via interpolation and standardized. The ECGs are labeled into 5 classes and only the first class, almost 60\% of the instances, contains normal heartbeats. The remaining classes are merged, making it a binary classification problem. $G_{\text{CF}}$ is trained on 4,500 instances and the remaining 500 ECGs are used to evaluate the counterfactual generator.

\subsubsection{Models}

$G_{\text{CF}}$ follows the adjusted RCGAN architecture described in section \ref{method:ts} and returns the counterfactual perturbations $\delta_{\text{CF}}$. The LSTM classifier reaches 99\% accuracy on the test set. The loss weights $w_{l_{1}}$, $w_{\text{CC}}$ and $w_{\text{G}}$ are unchanged from the image experiments and $w_{\text{M}}$ is kept at the default value of 1. We compare our RCGAN generator with CFProto, a counterfactual generation method guided by class-specific prototypes \citep{vanlooveren_cfproto}. More details can be found in \cref{sec:ts_details}.

\subsubsection{Evaluation}

\begin{figure}[t]
    \centering
    \includegraphics[width=0.99\columnwidth]{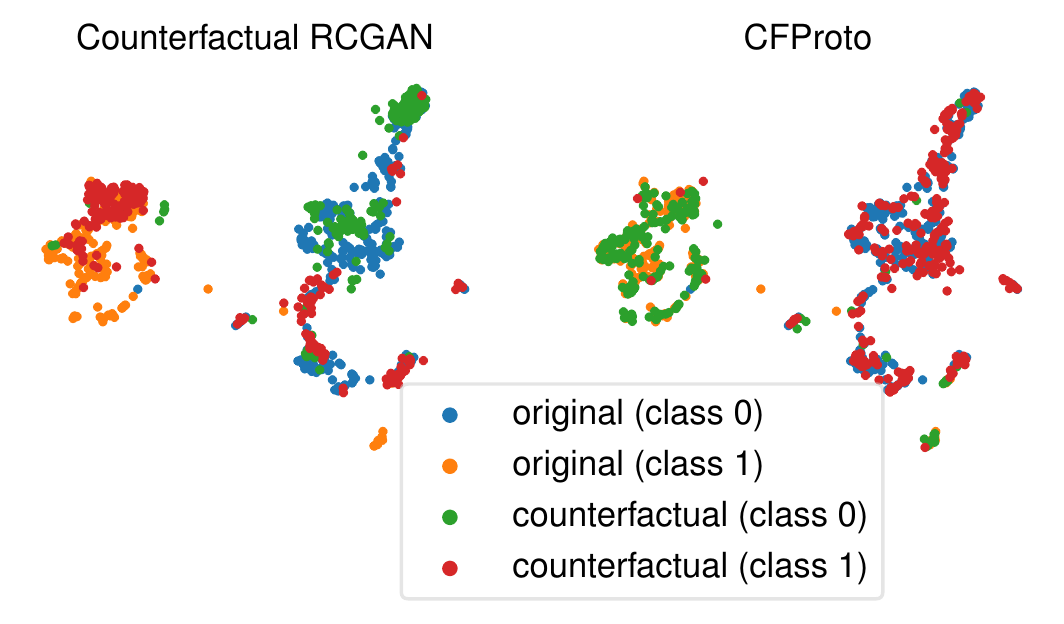}
    \caption{UMAP embeddings for the ECGs in the test set and the generated counterfactual instances using $G_{\text{CF}}$ and CFProto. Overall, $G_{\text{CF}}$ manages to flip the instances from class 0 well onto the distribution of class 1 and vice versa. CFProto also stays in-distribution but remains within the original class instead of the target class.}
    \label{fig:umap_ts_cf}
\end{figure}

\begin{figure}[ht]
    \centering
    \includegraphics[width=0.99\columnwidth]{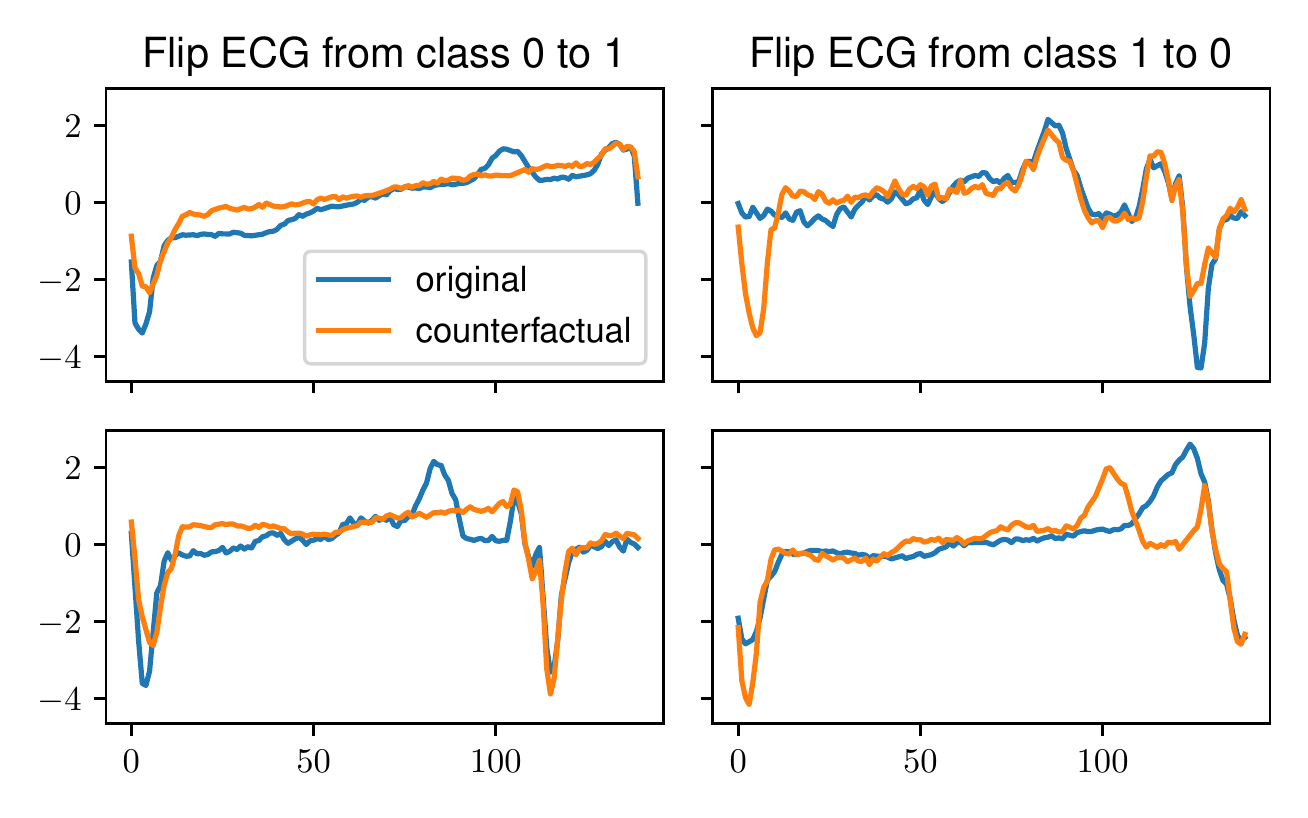}
    \caption{Counterfactual instances for ECGs in the test set. $G_{\text{CF}}$ flips the ECGs from the normal class 0 to class 1 in the first column and vice versa in the second column.}
    \label{fig:ts_cf}
\end{figure}

\begin{figure}[ht]
    \centering
    \includegraphics[width=0.99\columnwidth]{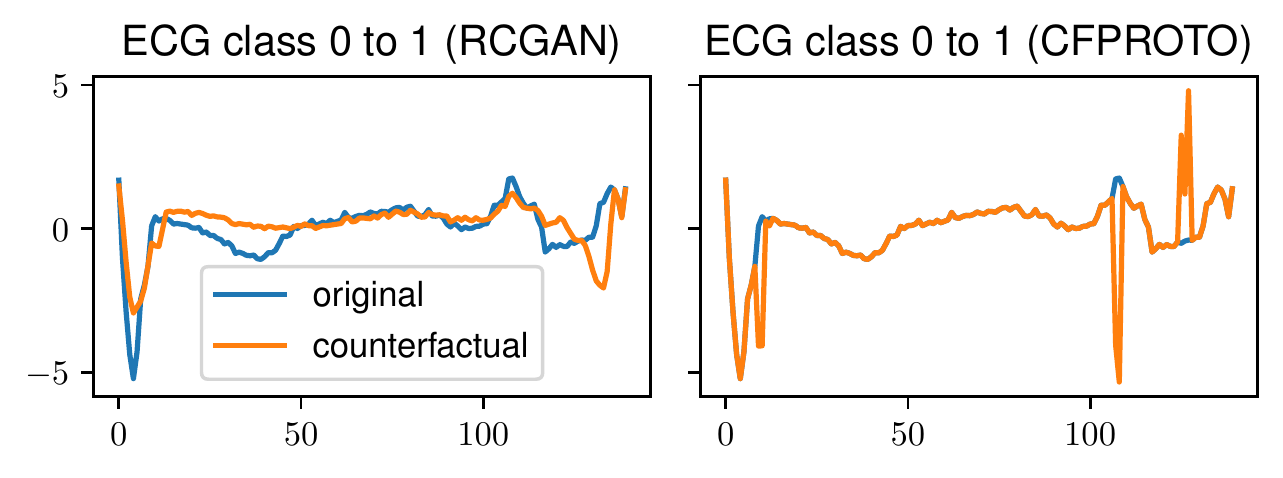}
    \caption{Counterfactuals generated for the same instance using $G_{\text{CF}}$ and RCGAN compared to CFProto. The counterfactual of CFProto is significantly sparser with $L_{1}$ of 0.23 for the perturbation compared to 0.48 for RCGAN, but is also out-of-distribution for class 1.}
    \label{fig:ts_cf_cfproto}
\end{figure}

The UMAP \citep{mcinnes_umap} embeddings of both the original test set instances and their counterfactuals in \Cref{fig:umap_ts_cf} illustrate that the perturbations $\delta_{\text{CF}}$ generated by $G_{\text{CF}}$ push the instance $x$ to the distribution of the counterfactual class. The counterfactuals $x_{\text{CF}}$ generated by CFProto on the other hand remain within the distribution of the original class. As a result, $\delta_{\text{CF}}$ applied by CFProto often resembles an adversarial perturbation instead of an in-distribution counterfactual explanation. These observations are supported by \Cref{tab:ts_comparison} and visualized in \Cref{fig:ts_cf}. The MMD$^{2}$ between the original test set and the counterfactuals $x_{\text{CF}}$, and the $L_{1}$ of the perturbations $\delta_{\text{CF}}$ are lower for CFProto than our proposed method. However, the tables turn when we look at the class-specific metrics. MMD$^{2}_{0}$ and MMD$^{2}_{1}$ are respectively the MMD$^{2}$ for instances $x$ or $x_{\text{CF}}$ belonging to classes 0 and 1 according to the model $M$. The class-specific MMD$^{2}$ values between the instances of the test set and generated counterfactuals strongly favour our method over the baseline. \Cref{fig:ts_cf_cfproto} illustrates how CFProto can generate more sparse but unrealistic counterfactuals which are out-of-distribution for the counterfactual class compared to $G_{\text{CF}}$. 

\begin{table}
\centering
\begin{tabular}{lcccc}
\toprule
Method  & MMD$^2$ & MMD$^2_{0}$ & MMD$^2_{1}$ & $L_{1}$ \\
\midrule
CFProto  &  \textbf{0.027}  & 0.22 & 0.21  &  \textbf{0.45}  \\
Ours    &  0.035 & \textbf{0.096} & \textbf{0.15} &  0.57  \\
\bottomrule
\end{tabular}
\caption{Method comparison between CFProto \protect\citep{vanlooveren_cfproto} and our counterfactual RCGAN approach for time series. MMD$^{2}_{c}$ represents the class-specific MMD$^{2}$ for instances of class $c$.}
\label{tab:ts_comparison}
\end{table}

\subsection{Tabular}
\subsubsection{Dataset}
We perform counterfactual search on the Adult Census dataset~\citep{uci}. The dataset consists of 32,561 rows of attributes of individuals together with a binary label indicating whether they earn below or over \$50K/p.a. Our pre-processed dataset consists of 12 features---8 categorical and 4 continuous. After one-hot encoding categorical features and performing mode-specific normalization~\citep{ctgan2019} this results in feature vectors of length 85. The classifier as well as the counterfactual generator are trained on 80\% of the data while counterfactual instances are generated and evaluated on the remaining 20\%.

\subsubsection{Models}
For the classifier, we use a 2-layer fully connected network which reaches 86\% accuracy on the test set. For the generative model, we use a conditional GAN as described in \Cref{method:tabular}. We set the relative loss weights to $w_G=1$, $w_M=1$ and $w_{l_1}=10$. Full details of the architectures and training procedures are available in \cref{sec:tabular_details}. For baselines we also generate counterfactual examples from two popular methods on tabular data---DiCE~\citep{mothilal_dice} and prototype counterfactuals~\citep{vanlooveren_cfproto}.

\subsubsection{Evaluation}

We check that our method can generate counterfactual instances across the whole range of the target distribution $y_{\text{T}}$ (see \cref{sec:tabular_details}). We compare our method with the baselines to gauge the in-distribution quality of the generated samples.

\begin{figure}[ht]
    \centering
    \includegraphics[width=0.95\columnwidth]{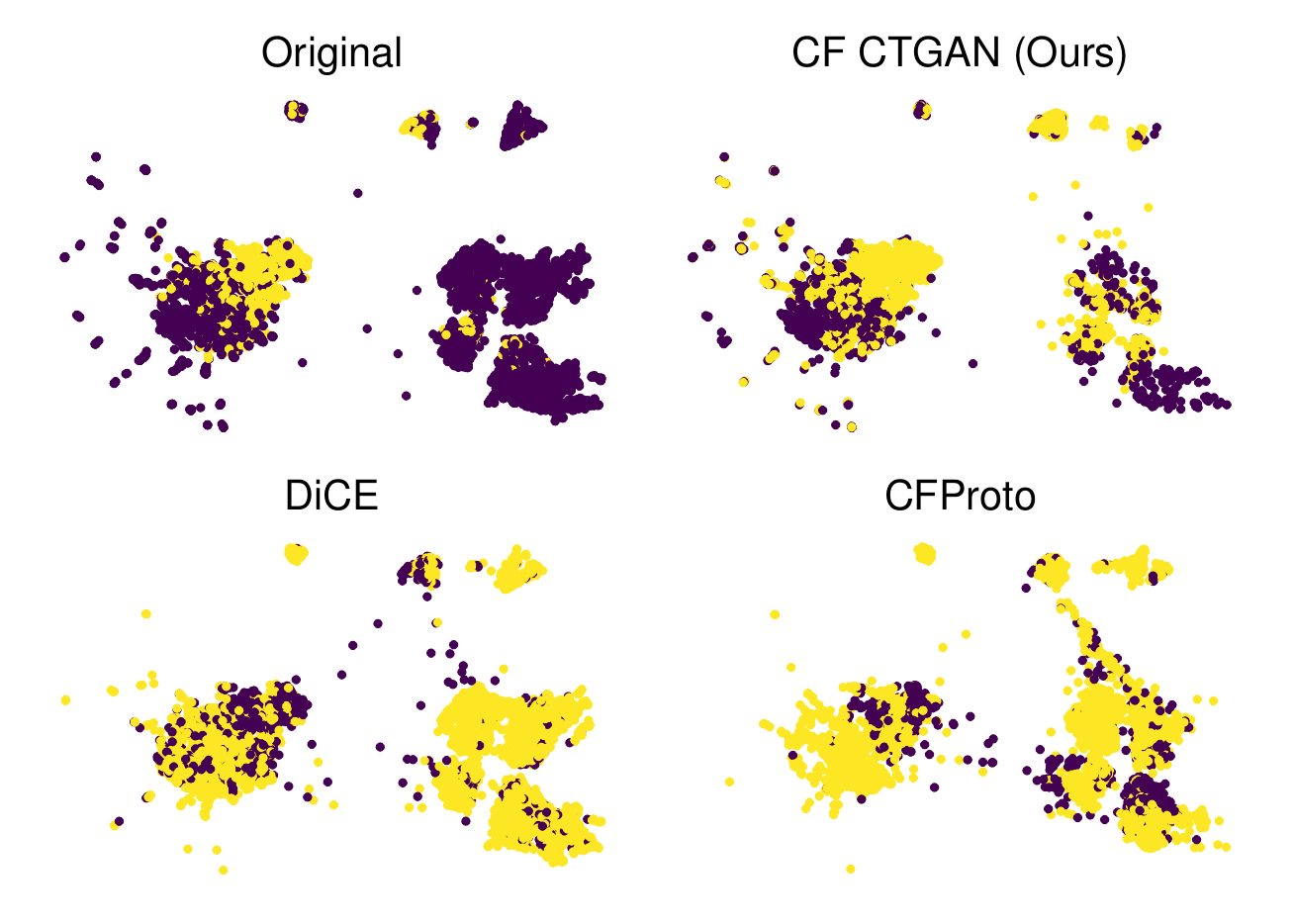}
    \caption{UMAP embeddings of the Adult dataset (Original) compared to the embeddings of the generated counterfactuals for each method. Color indicates predicted class by the classifier.}
    \label{fig:umap_tabular}
\end{figure}

\Cref{fig:umap_tabular} shows the UMAP embeddings of the original test set as well as the embeddings of counterfactuals generated for each method (one per instance in the test set). Whilst on the whole the overall data distribution is preserved by all three methods, the class-specific distributions of the generated counterfactuals are not well modelled by DiCE or CFProto. The class-specific distributions of DiCE and CFProto counterfactuals suggest that the methods favour sparsity over class realism. On the other hand, the GAN method is more successful in preserving the class-conditional distributions.

This qualitative view is confirmed by calculating the MMD distance between the set of counterfactual instances and the original instances. \Cref{tab:tabular_comparison} shows that our method generates the most in-distribution (both overall and class-conditional) counterfactual instances, Additionally, we include the average $L_1$ and $L_0$ distances between the set of counterfactuals and the original instances. $L_1$ is reported on the standardized numerical columns whilst $L_0$ is reported on the categorical columns, effectively measuring the average number of categories changed by going from the original instance to a counterfactual one. We can see that different methods prioritize sparsity on numerical and categorical columns differently. We also note that there is always a trade-off between the sparsity and realism of the generated counterfactuals. All methods provide some degree of customizing these trade-offs depending on the desired properties of counterfactuals.

\begin{table}[t]
\centering
\begin{tabular}{lccccc}
\toprule
Method  & MMD$^2$ & MMD$_0^2$ & MMD$_1^2$ & $L_1$ & $L_0$ \\
\midrule
DiCE & 0.047 & 0.030 & 0.090 & 0.311 & \textbf{0.425} \\
CFProto & 0.059  & 0.054 & 0.092 & 0.124 & 4.95 \\
Ours    & \textbf{0.021} & \textbf{0.023} & \textbf{0.015} & \textbf{0.065} & 2.13   \\
\bottomrule
\end{tabular}
\caption{Method comparison between DiCE \protect\citep{mothilal_dice}, CFProto \protect\citep{vanlooveren_cfproto} and our counterfactual CTGAN approach for tabular data. MMD$^{2}_{c}$ represents the class-specific MMD$^{2}$ for instances of class $c$.}
\label{tab:tabular_comparison}
\end{table}

\Cref{fig:cf_all_modality} shows two counterfactual instances where the original instances were predicted class 0 (<\$50K) and class 1 (>\$50K) respectively (more examples in \cref{sec:tabular_samples}). We can see that in both cases the counterfactual generator has focused on the features ``Occupation", ``Capital Gain" and ``Hours per week" to change the prediction to the opposite class.

\section{Conclusion}
In this paper we introduce a flexible, modality agnostic framework to generate counterfactual explanations. We show on image, time series and mixed type tabular datasets that the method is able to create batches of in-distribution, sparse counterfactual instances which match the prediction target with a single forward pass of a conditional generative model. The method can be used for various predictive tasks such as classification or regression. 


\bibliographystyle{named}
\bibliography{main}

\clearpage
\appendix

\section{Experiment Details}
\subsection{Image}\label{sec:image_details}

We follow the BigGAN architecture with 5 ResBlocks in the generator $G_{\text{CF}}$ and 6 in $D$, as suggested for 128x128 resolution images \citep{brock_biggan}. To reduce the memory footprint, we reduce the channel multiplier from 64 to 24 and do not include self-attention blocks \citep{zhang_sagan}. We also do not apply orthogonal regularization \citep{brock_reg}. Since $G_{\text{CF}}$ takes the original instance $x$ directly as the input of the generator instead of a random noise vector $z$, no upsampling takes place in the ResBlocks. A 20-dimensional normally distributed noise vector is injected in each ResBlock via the class-conditional BatchNorm \citep{dumoulin_ccbn} layers. The BatchNorm layers in $G_{\text{CF}}$ are further conditioned by separate 64-dimensional embedding vectors for $y_{\text{M}}$ and $y_{\text{T}}$. $y_{\text{M}}$ and $y_{\text{T}}$ use separate embedding layers and do not share weights. $G_{\text{CF}}$ directly models $x_{\text{CF}}$. $D$ is conditioned via a projection head \citep{miyato_projhead} on $y_{\text{M}}$ for the real instances $x$ and on the target predictions $y_{\text{T}}$ for the counterfactuals $x_{\text{CF}}$. Similar to the original BigGAN paper, we use Adam optimizers \citep{kingma_adam} for both $D$ and $G_{\text{CF}}$ with learning rates of respectively 2e-4 and 5e-5. We apply 8 steps of gradient accumulation for $G_{\text{CF}}$ and 16 for $D$. We train for a total of just over 60,000 steps with batch size 20. While this is significantly shorter and with smaller batch size than an original BigGAN model trained on CelebA for 400,000 training steps and batch size 50, it reaches a similar FID (5.76 vs. 4.54) and slightly higher IS (3.33 vs. 3.23) score \citep{schonfeld_ugan}. For sampling, we use an EMA of the weights of $G_{\text{CF}}$ with a weight decay of 0.9999.

The ResNet-18 \citep{he_resnet} classifier is trained for 10 epochs on the CelebA train set and achieves 81.8\% accuracy on the test set. The Adam optimizer is used with a learning rate of 1e-3.

The CelebA instances are scaled between -1 and 1, and random horizontal flips are applied as data augmentation. The training set is imbalanced and contains 59,765, 67,023, 18,315 and 17,667 for the respective classes \textit{young/smiling, young/non-smiling, old/smiling, old/non-smiling}. The target transformation prediction function $T$ flips the predictions between classes as one-hot encodings but also allows for soft targets.

\begin{figure}[ht]
    \centering
    \includegraphics[width=0.95\columnwidth]{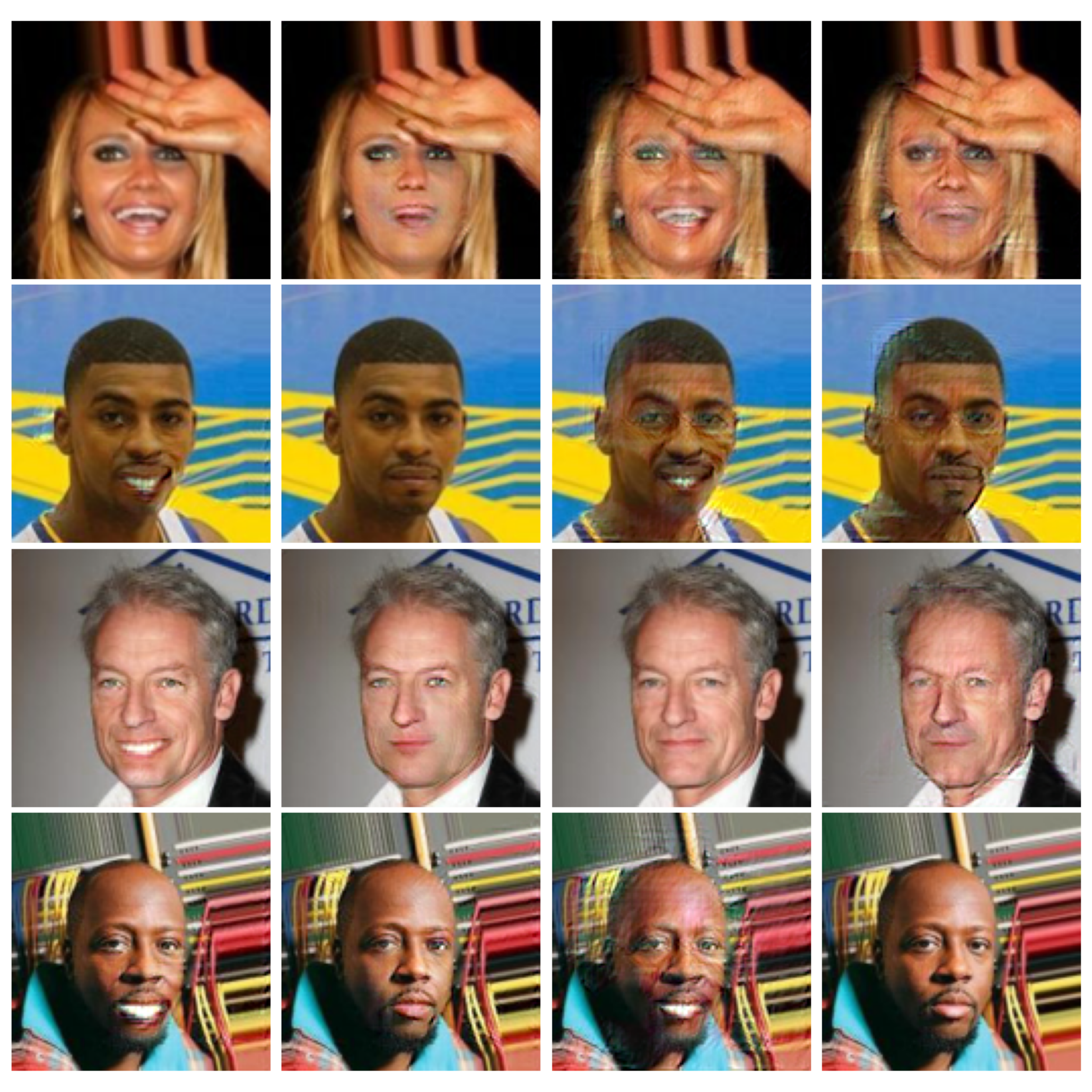}
    \caption{Counterfactual instances generated by BIN for each of the classes in the test set of CelebA. From left to right: young/smiling, young/non-smiling, old/smiling, old/non-smiling. The images on the diagonal are the original instances and the off-diagonal elements are counterfactuals gen.}
    \label{fig:img_bin}
\end{figure}

We compare our framework against the Born Identity Network (BIN) framework of \cite{oh_bornid}. A BIN is also a GAN trained to generate counterfactual instances, however the generator instead has a U-Net encoder-decoder structure whereby the encoder adopts the convolutional base of the model being explained and each block in the decoder, which is skip-connected to a corresponding block in the encoder, is additionally conditioned on the class being targeted. The convolutional base of the ResNet-18 classifier is frozen and used as the encoder in the U-Net. Only the U-Net's decoder, which consists of upsampling ResNet blocks, is randomly initialized and trained in the GAN setting. The U-Net is then used to generate the counterfactuals. The GAN discriminator also adopts the ResNet-18 architecture but is initialized with pretrained ImageNet weights. We found it necessary to include L2 regularisation of the discriminator's weights in order to provide training signal to the generator (with associated weight $1e-4$) and trained both the generator and discriminator using the Adam optimizer, finding the best learning rate to be $1e-3$. We saved the model after each epoch and retained the version that produced the most realistic looking counterfactuals on the validation set. Figure~\ref{fig:img_bin} presents counterfactuals generated by this model for instances in the test set.

The dimensionality reduction step applied to the original and counterfactual test set images is done with a randomly initialized convolutional encoder \citep{rabanser_failingloudly} which projects the images on a 32-dimensional encoding vector. The MMD$^2$ is then computed on the image encodings. The encoder consists of 3 convolution layers with kernel size 7, stride 2 and respectively 64, 128 and 256 filters. Each layer is followed by a Leaky ReLU \citep{maas_lrelu} activation with a negative slope of 0.01. Finally the output is flattened and fed into a linear layer which projects the instance on the encoding dimension.

\subsection{Time Series}\label{sec:ts_details}

$G_{\text{CF}}$ and $D$ consist of LSTMs with a hidden size of 512 as well as a linear output layer to predict respectively the counterfactual perturbation $\delta_{\text{CF}}$ and the discriminator logits. $G_{\text{CF}}$ takes as input at each step $n$ $x_{n}$, a 64-dimensional normally distributed noise vector $z$ and 64-dimensional embedding vectors for both $y_{\text{M}}$ and $y_{\text{T}}$. $z$ is independently sampled for each step $n$ in the sequence. Similar to the image experiments, the embedding layers are not shared between $y_{\text{M}}$ and $y_{\text{T}}$. Both $G_{\text{CF}}$ and $D$ use Adam optimizers with a learning rate of 1e-3.
We train for a total of almost 25,000 steps with batch size 32.

The binary LSTM classifier is bidirectional with a hidden size of 256. The classifier is trained for 5 epochs with an Adam optimizer with learning rate 1e-3 and reaches a test set accuracy of 99\%.

Each ECG has a sequence length of 140 and is standardized using the training set mean and standard deviation. Classes 2 to 5 of the original data set are merged into 1 class to make it a binary classification problem. The data set would otherwise be extremely imbalanced leading to varying counterfactual performance for different classes. The target transformation prediction function $T$ flips the predictions between classes as one-hot encodings but also allows for soft targets.

\subsection{Tabular}\label{sec:tabular_details}
We pre-process the Adult dataset to contain 12 features---8 categorical (Workclass, Education, Marital Status, Occupation, Relationship, Race, Sex, Country) and 4 real-valued (Age, Capital Gain, Capital Loss, Hours per week). We apply one-hot encoding to the categorical variables and mode-specific normalization~\citep{ctgan2019} to the numerical variables resulting in feature vectors of length 85.

The $G_\text{CF}$ and $D$ architectures are directly reused as-is from \citet{ctgan2019}. We train $G_\text{CF}$ and $D$ adversarially with a hinge loss, both the generator and discriminator use an Adam optimizer with a learning rate of 2e-4. The networks are trained for 1000 epochs with a batch size of 512.

The binary classifier is a 2-layer fully-connected network with 40 neurons in each hidden layer and ReLU activations. The classifier is trained for 5 epochs with an Adam optimizer with learning rate 1e-3 and reaches a test set accuracy of 86\%.

\begin{figure}[ht]
    \centering
    \includegraphics[width=0.95\columnwidth]{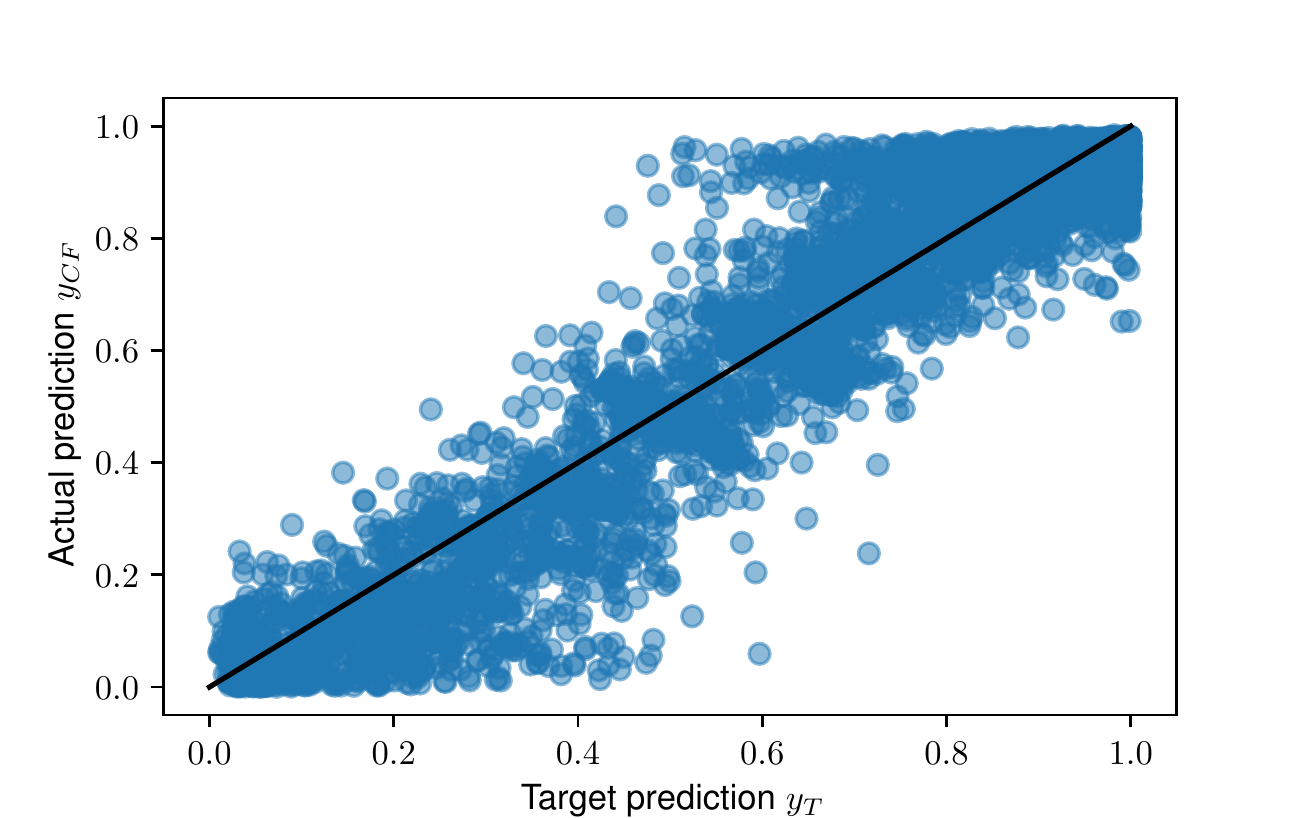}
    \caption{Scatter plot of the target predictions on class 0 vs the actual prediction on the counterfactual instances on the test set.}
    \label{fig:tabular-scatter}
\end{figure}

\Cref{fig:tabular-scatter} shows a scatter plot of target predictions $y_\text{T}$ (for class 0) and actual predictions on the counterfactual instances $y_{\text{CF}}$ across the test set. During testing the target prediction distribution for each instance in the test set was set to be the opposite to the prediction distribution on the original instance (i.e. $y_\text{T}=1-y_{\text{M}}$).

\section{Generating Counterfactuals with Autoencoders}\label{sec:cf_ae}

Our counterfactual generation framework is agnostic to the generative modeling paradigm and not restricted to GANs. Here, we consider an alternative approach to create counterfactuals for ECGs based on autoencoders. An auto-encoder consists of an encoder network $E_{\theta}$ and a decoder network $D_{\theta^\prime}$. The encoder takes the original instance as input and returns a hidden vector $h$. The input of the decoder consists of $h$, the classifier predictions $y_{\text{M}}$ and the target predictions $y_{\text{T}}$. The goal of $D_{\theta^\prime}$ is then to model the counterfactual perturbation $\delta_{\text{CF}}$. The obtained counterfactual instance can be represented as follows:

\begin{equation}\label{eq:x_cf_ae1}
    \begin{aligned}
        x_{\text{CF}} &= x + D(E(x), y_{\text{M}}, y_{\text{T}}).
    \end{aligned}
\end{equation}

While the prediction and sparsity loss terms $L_{\text{M}}$ and $L_{l_{1}}$ can remain unchanged from the GAN setting, we still need to encourage $x_{\text{CF}}$ to stay within the training data distribution since we cannot leverage the GAN's discriminator to improve the realism of the counterfactual instance. Instead, we introduce a loss term which minimizes the MMD \citep{gretton_mmd} between the counterfactuals $x_{\text{CF}}$ and training instances $x_{\in T}$ which also belong to the target class $T$ according to the model $M$:

\begin{equation}\label{eq:x_cf_ae2}
    \begin{aligned}
        L_{\chi} &= w_{\chi} \text{MMD}(x_{\text{CF}}, x_{\in T}) \\
        L_{G_{\text{CF}}} &= L_{\text{M}} + L_{l_{1}} + L_{\chi}. \\
    \end{aligned}
\end{equation}

Note that we still only rely on model predictions and do not require access to ground truth labels.

\section{Samples}

\subsection{Image}\label{sec:image_samples}

\begin{figure}[ht]
    \centering
    \includegraphics[width=0.95\columnwidth]{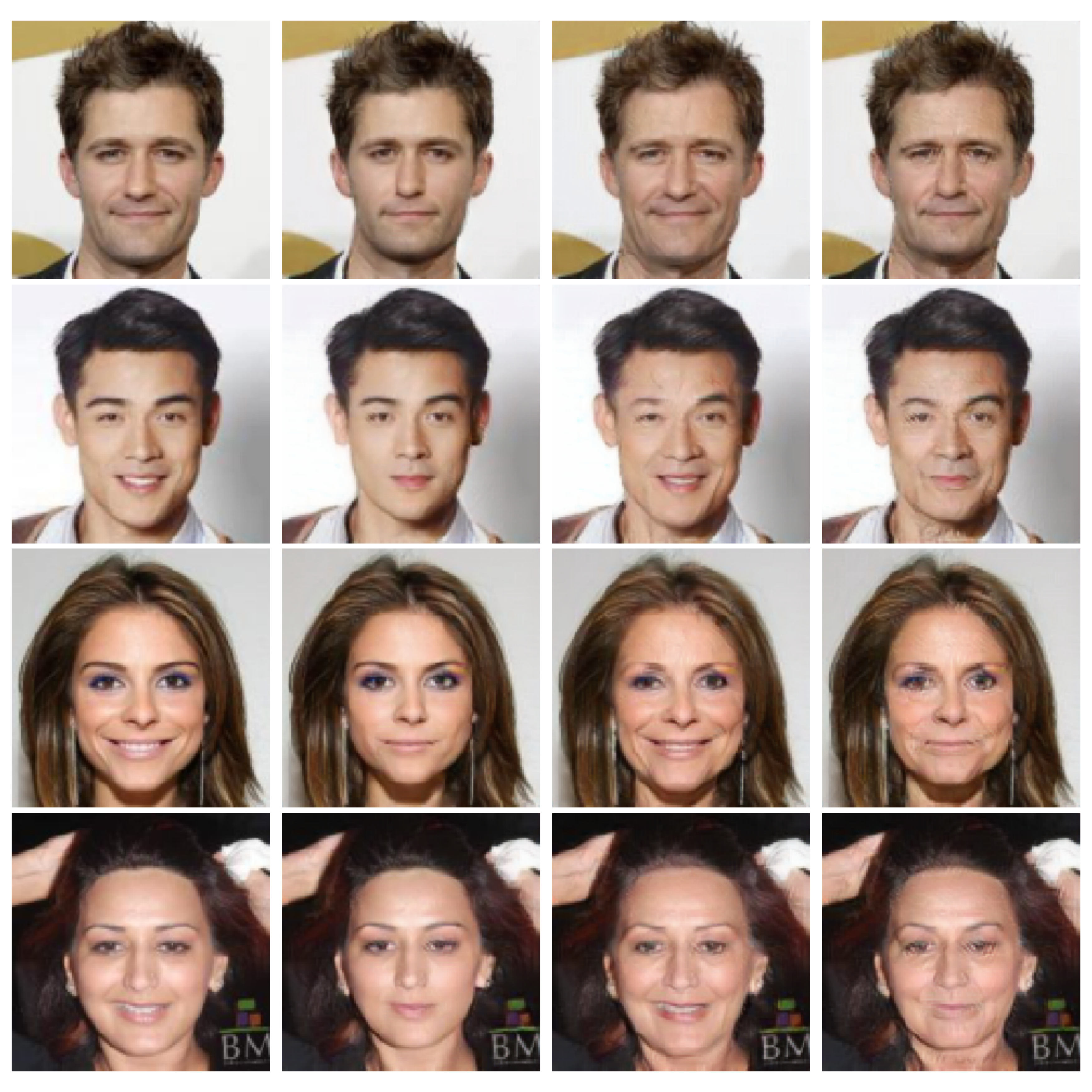}
    \caption{Counterfactual instances for class 0 (young/smiling) in the test set of CelebA. The images in the first column are the original instances.}
    \label{fig:img_cf_sm0}
\end{figure}

\begin{figure}[ht]
    \centering
    \includegraphics[width=0.95\columnwidth]{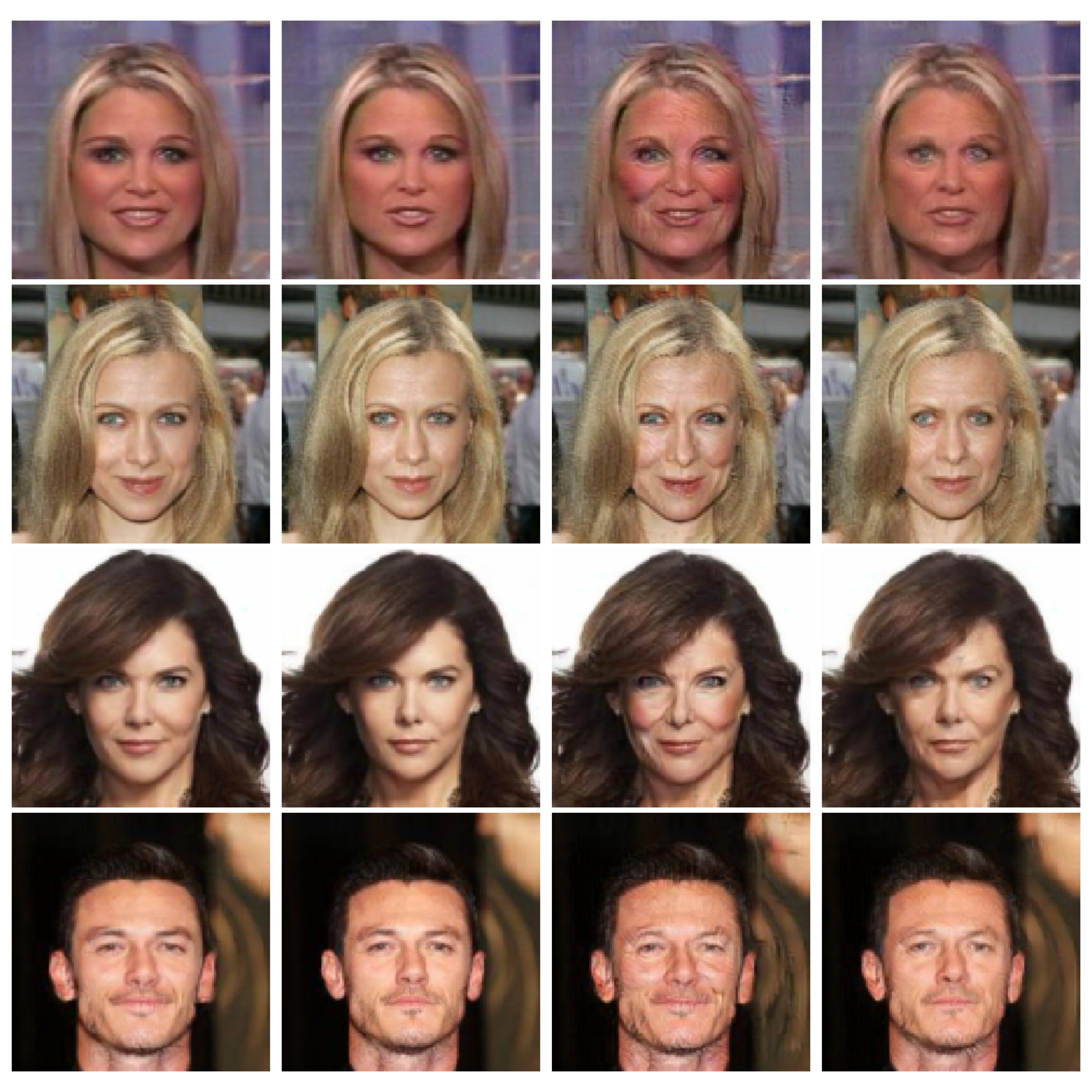}
    \caption{Counterfactual instances for class 1 (young/non-smiling) in the test set of CelebA. The images in the second column are the original instances.}
    \label{fig:img_cf_sm1}
\end{figure}

\begin{figure}[ht]
    \centering
    \includegraphics[width=0.95\columnwidth]{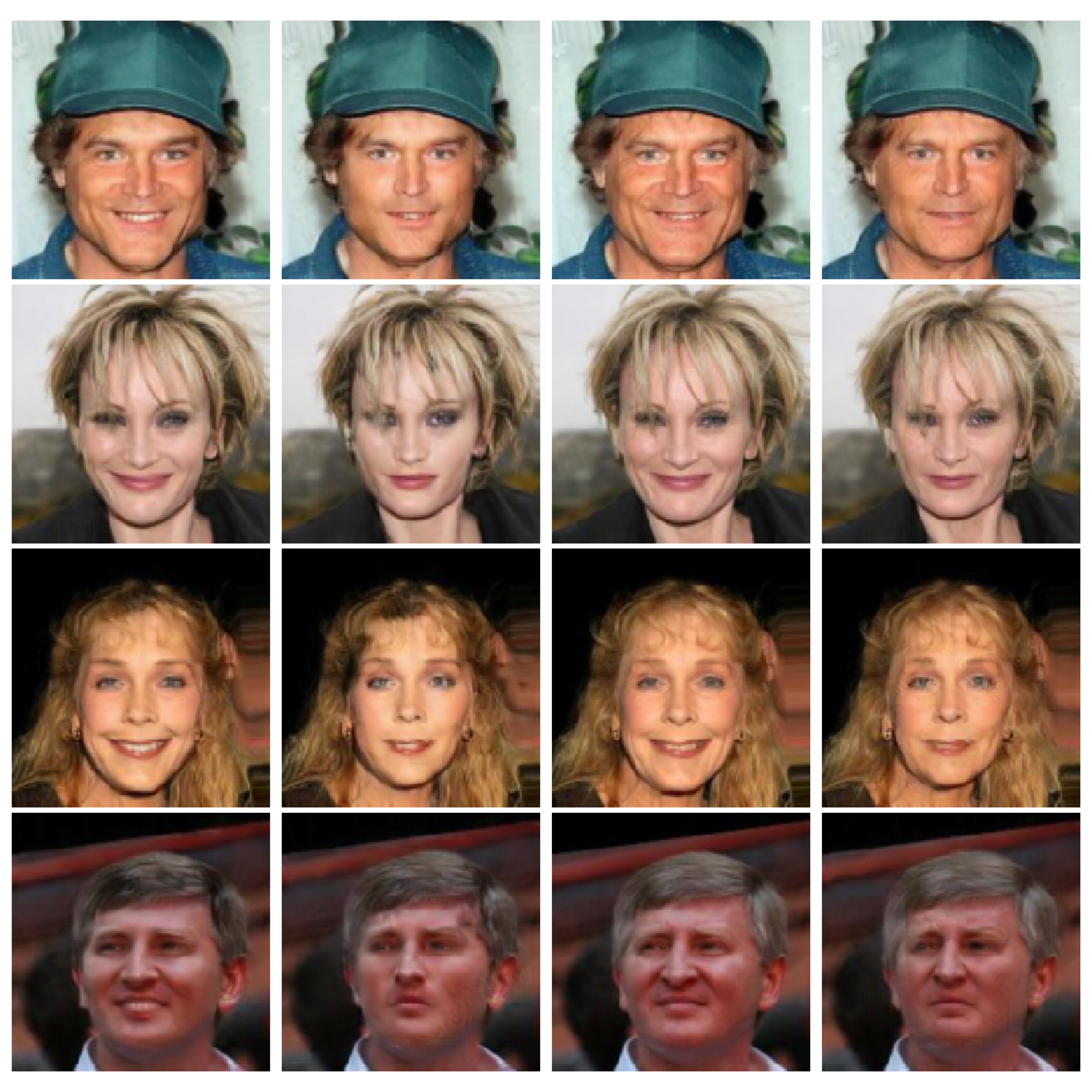}
    \caption{Counterfactual instances for class 2 (old/smiling) in the test set of CelebA. The images in the third column are the original instances.}
    \label{fig:img_cf_sm2}
\end{figure}

\begin{figure}[ht]
    \centering
    \includegraphics[width=0.95\columnwidth]{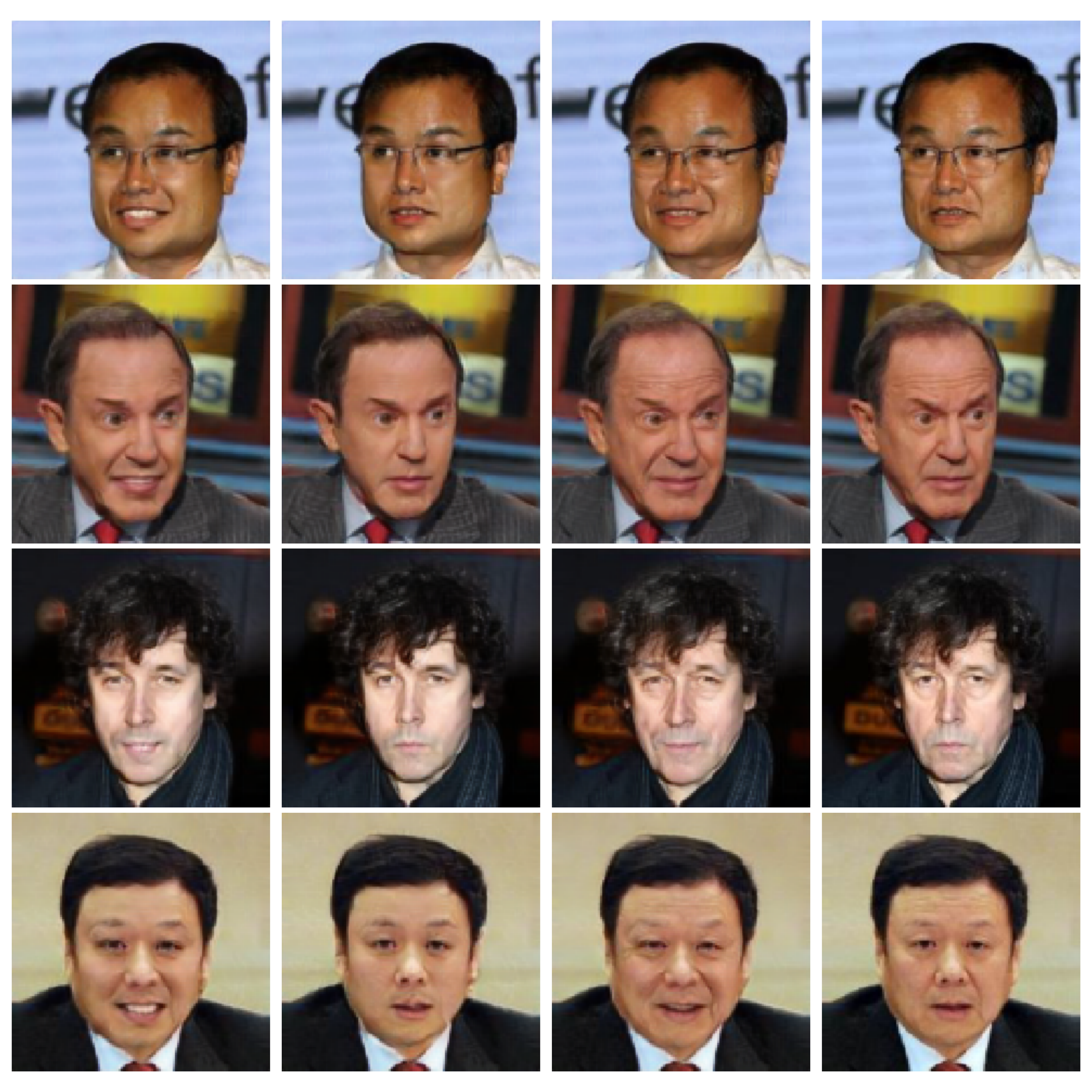}
    \caption{Counterfactual instances for class 3 (old/non-smiling) in the test set of CelebA. The images in the last column are the original instances.}
    \label{fig:img_cf_sm3}
\end{figure}

\clearpage
\subsection{Time Series}\label{sec:ts_samples}

\begin{figure}[ht]
    \centering
    \includegraphics[width=0.95\columnwidth]{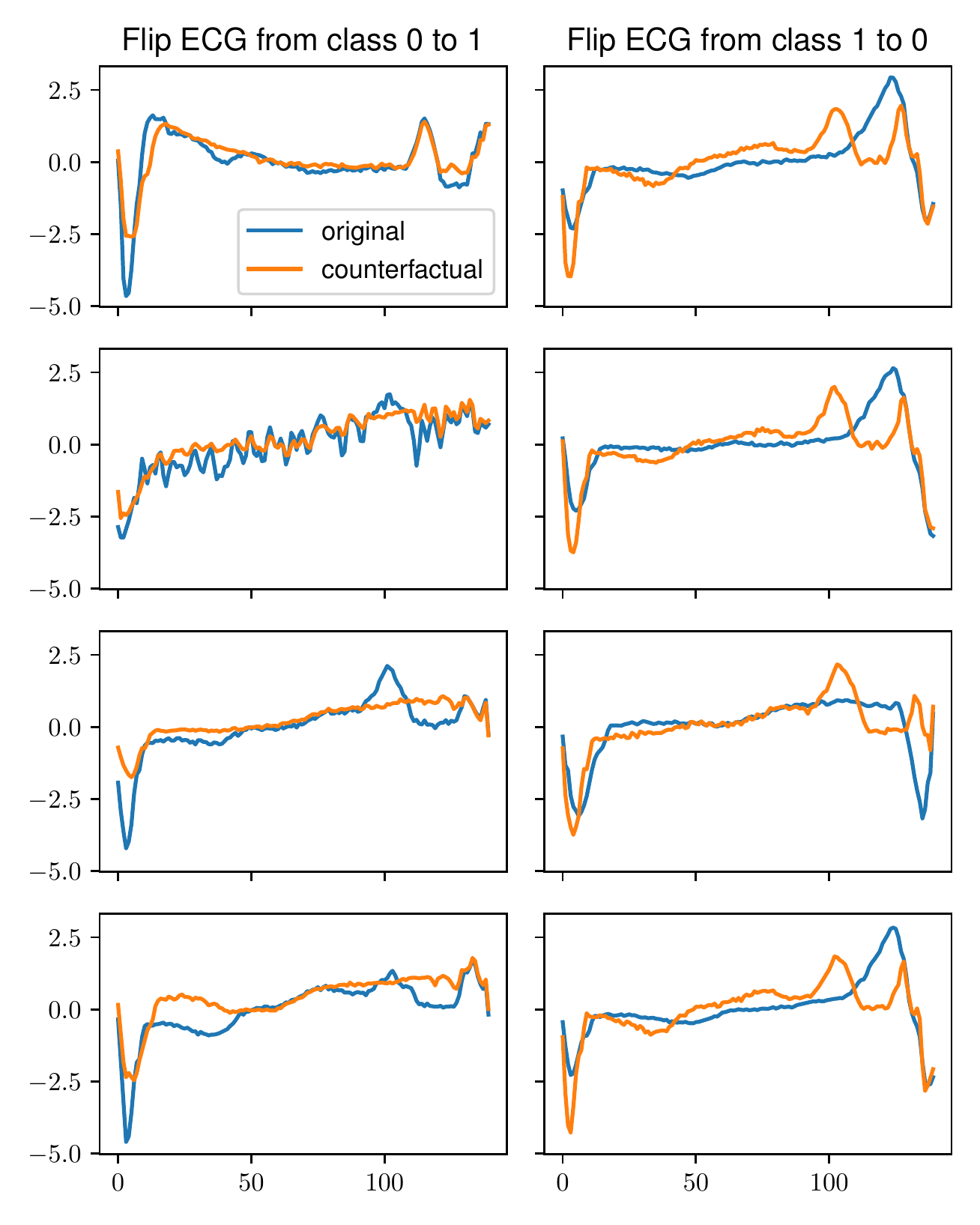}
    \caption{Counterfactual instances for ECGs in the test set. $G_{\text{CF}}$ flips the ECGs from the normal class 0 to class 1 in the first column and vice versa in the second column.}
    \label{fig:img_ts_sm}
\end{figure}

\clearpage
\subsection{Tabular}\label{sec:tabular_samples}

\begin{figure}[ht]
    \centering
    \includegraphics[width=0.92\columnwidth]{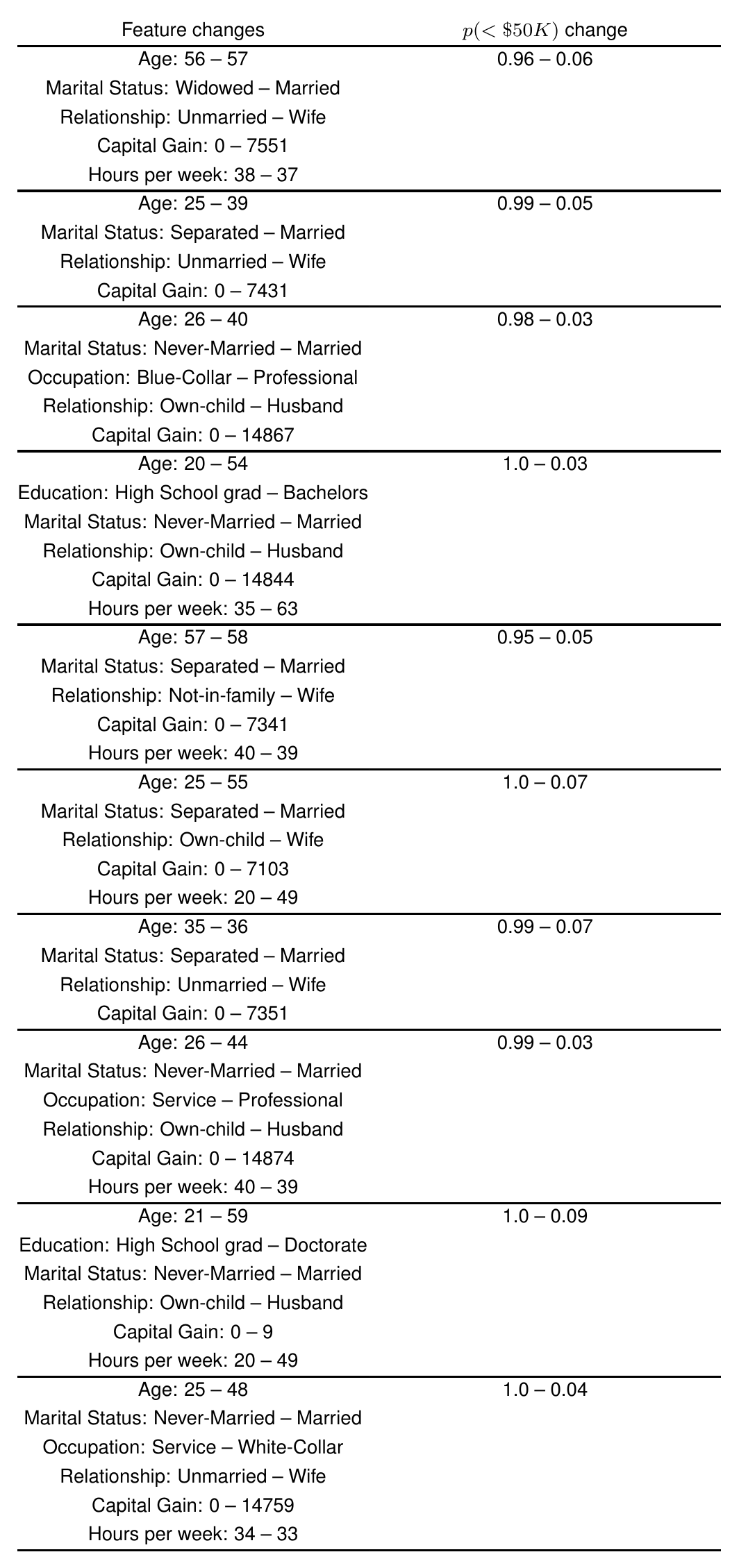}
    \caption{Counterfactual instances generated by CF-CTGAN for class 0 ($<\$50K$) in the Adult dataset.}
    \label{fig:img_tab_sm1}
\end{figure}

\newpage
\begin{figure}[ht]
    \centering
    \includegraphics[width=0.92\columnwidth]{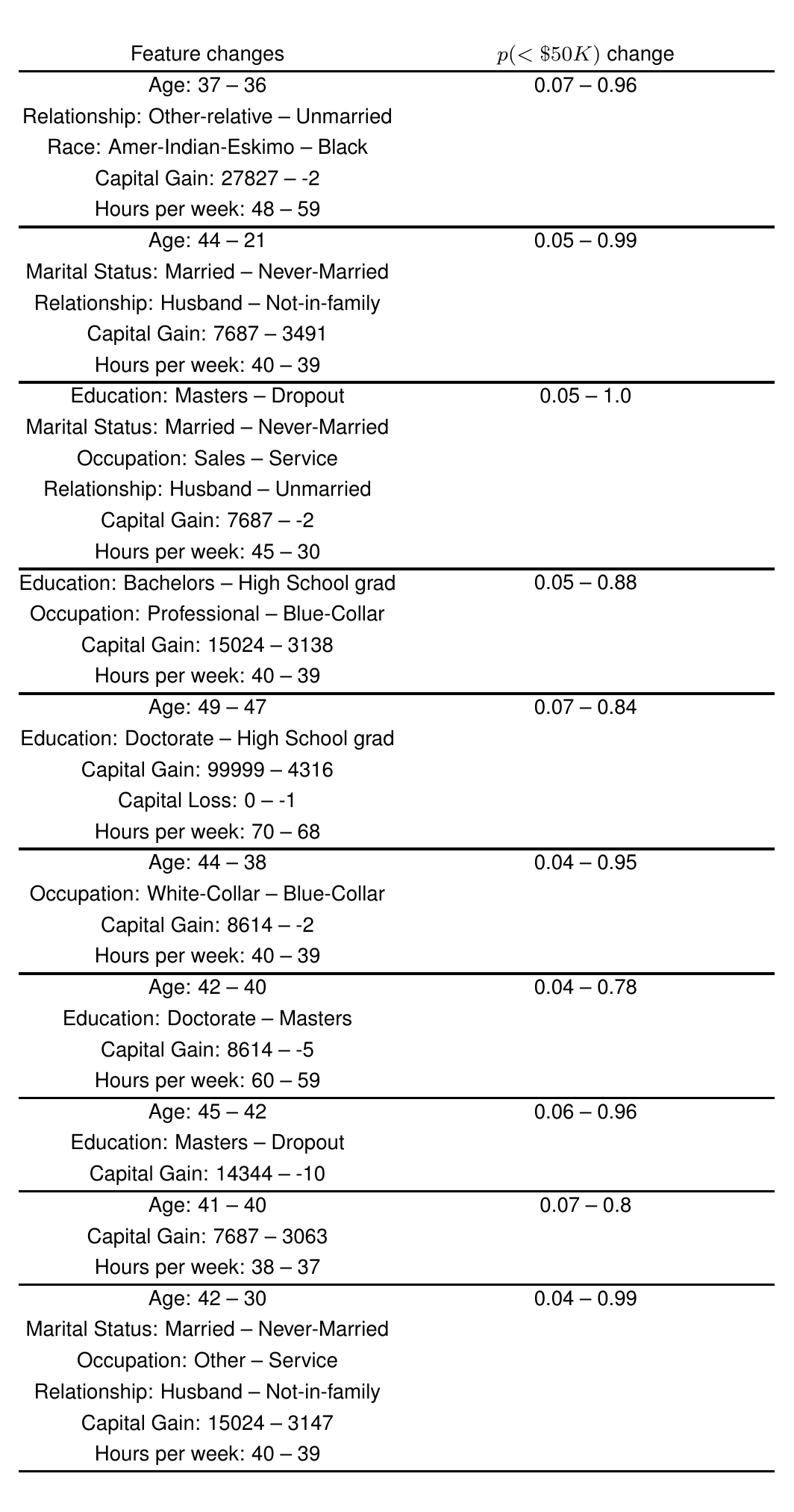}
    \caption{Counterfactual instances generated by CF-CTGAN for class 1 ($>\$50K$) in the Adult dataset.}
    \label{fig:img_tab_sm2}
    
\end{figure}

\end{document}